\documentclass[11pt]{article}

\usepackage[]{acl}                
\usepackage{times}
\usepackage{latexsym}
\usepackage[T1]{fontenc}
\usepackage[utf8]{inputenc}
\usepackage{microtype}
\usepackage{inconsolata}

\usepackage{booktabs}
\usepackage{graphicx}
\usepackage{amsmath}
\usepackage{amssymb}

\PassOptionsToPackage{hyphens}{url}
\usepackage{url}
\newcommand{\mv}{\texttt{math-verify}}
\newcommand{\mvl}{\texttt{mv-latex}}
\newcommand{\mve}{\texttt{mv-expr}}
\newcommand{\strips}{\texttt{strip-string}}
\newcommand{\sympyc}{\texttt{sympy-cascade}}

\title{Where the Verifier Fails: A Category-Level Audit of
       Reward Signals in RLVR}

\author{
  Esther Xin \\
  Independent Researcher \\
  \texttt{estherxin0011@gmail.com} \\
  {\small Code and data: \url{https://github.com/ethxin0011/verifier-error-budget}}
}

\begin{document}
\maketitle

\begin{abstract}
Reinforcement learning with verifiable rewards (RLVR) and standard
benchmark evaluation both rely on an automatic verifier that turns a
free-text answer into a binary reward. Prior work reports that one
evaluation harness accepts only about 94\% of its own ground-truth
answers, blaming LaTeX parsing. That is an aggregate: it does not say
which answer forms consume the error budget.

We supply the decomposition. We apply metamorphic testing to the verifier
rather than the model, generating \emph{certified-equivalent} answer
variants---rewrites that preserve mathematical meaning by construction, so
any rejection is a provable false negative needing no human
adjudication---and measure rejection per answer category across four
widely used verifiers over 307{,}420 verdicts.

We find three things. \textbf{(1)} Self-validation ranges from 53.8\% to
95.2\% on identical inputs, a 41.3-point spread. The published figure
describes one implementation, not the task; two configurations of the
same library disagree on 49.9\% of pairs. \textbf{(2)} The residual is not
spread across parsing categories but concentrated in whitespace and
punctuation, which account for 93.0\% of in-contract failures for the
default LaTeX configuration. A trailing period or newline dominates the
budget. \textbf{(3)} Separating rejection from execution failure shows
that verifiers with similar aggregate error fail for opposite reasons, and
that a reference numeric cascade accepts off-by-one wrong answers as a
step function of magnitude---0\% below $10^4$, 100\% at or above---because
its relative tolerance is scale-invariant.
\end{abstract}

\section{Introduction}
\label{sec:intro}

Reinforcement learning with verifiable rewards (RLVR) has become the
dominant post-training recipe for mathematical reasoning. Its central
appeal is that the reward requires no human annotation
\citep{guo2025deepseekr1}: a model emits an answer, an automatic verifier
checks it against a reference, and the resulting binary signal drives
policy optimisation. The same verifiers serve a second role, converting
free-text model outputs into the accuracy figures reported on benchmarks
such as MATH and GSM8K \citep{hendrycks2021math,cobbe2021gsm8k}.

This shifts a substantial burden onto a component that is rarely
examined. The verifier is not a mathematical oracle; it is a program that
extracts a substring, normalises it, and compares it to a reference under
some notion of equivalence. Every step is a design decision, and every
decision has a failure mode. If the verifier rejects a correct answer,
the policy is penalised for behaviour it should be rewarded for. If it
accepts an incorrect one, the policy is rewarded for being wrong.

Recent work has begun to treat this seriously.
\citet{cai2025noisy} formalise verifier unreliability as a stochastic
reward channel with asymmetric noise rates $\rho_0$ (false positive) and
$\rho_1$ (false negative), and derive backward and forward corrections to
the policy gradient; notably, the forward correction requires only the
false-negative rate as input. \citet{egashira2026delay} show that this
framing is incomplete in an important way: verifier errors in practice are
\emph{systematic} rather than random, and systematic false positives can
drive outcomes ranging from suboptimal plateaus to collapse, with the
result determined by the \emph{pattern} of errors rather than the
aggregate rate. \citet{huang2025accuracy} document that rule-based
checkers and model-based judges fail in opposite directions, the former
through brittle parsing and the latter through reward hacking.

Both threads point at the same missing quantity. Corrections need error
rates as parameters; the systematic-error analysis needs the \emph{shape}
of the error, not merely its magnitude. Yet the reliability literature
reports aggregates. The most concrete published figure is that a standard
evaluation harness validates ground truth against itself at roughly 94\%
accuracy, with the residual attributed to LaTeX parsing failures
\citep{hfguidebook}. That number tells a practitioner that six percent of
correct answers are lost somewhere. It does not say where.

\paragraph{This paper supplies the decomposition.}
We adapt metamorphic testing
\citep{asgari2025mtsurvey,hyun2024metal}, a technique normally applied to
models, and point it at the verifier instead. Given a gold answer $g$ and
a transformation $T$ that preserves mathematical meaning \emph{by
construction}, the pair $(g, T(g))$ has a known correct verdict. Any
verifier rejecting it commits a false negative that is certified: no
human adjudication, no trusted arbiter, no appeal to a stronger model. The
dual construction, applying a meaning-\emph{changing} transformation and
observing acceptance, certifies false positives. Because ground truth
comes from the construction rather than from labelling, the procedure
scales combinatorially in golds $\times$ transforms $\times$ verifiers and
runs entirely on CPU.

We apply this to four widely used implementations across 43
transformations in 14 strata, yielding 307{,}420 verdicts. A contract
matrix records which strata each verifier claims to handle, so that a
normalizer which never advertised answer extraction is not scored as
defective for rejecting a boxed answer; out-of-contract behaviour is
reported but never counted as a bug. We further separate \emph{rejection}
from \emph{execution failure}, so that a verifier which crashes on an
input cannot be credited with a correct verdict.

\paragraph{Findings.}
\textbf{(1) Self-validation is implementation-specific, not a property of
the task.} Acceptance of certified-equivalent variants ranges from 53.8\%
to 95.2\% across the four implementations on identical inputs, a spread of
41.3 points. Two configurations of the \emph{same library} differ by 26
points and disagree on 49.9\% of individual pairs. A benchmark accuracy
figure is therefore a function of the verifier configuration, which is
almost never reported.

\textbf{(2) The residual is string hygiene, not grammar.} Whitespace and
punctuation handling accounts for 93.0\% of in-contract failures for the
default LaTeX configuration and 74.1\% for the plain-expression
configuration. A trailing period or a newline, which no reasonable
specification treats as semantically meaningful, dominates the error
budget. This reframes the problem: the prior attribution to LaTeX parsing
suggests difficult grammar work, whereas the measured distribution points
at normalisation that is largely trivial to fix.

\textbf{(3) Similar aggregates hide opposite failure modes.} Separating
rejection from execution failure shows that \sympyc{} and \strips{} have
comparable aggregate residuals (12.7\% and 4.8\%) arising from disjoint
causes: \sympyc{} returns a verdict on only 87.3\% of inputs but is
correct on \emph{every} input it judges, whereas \strips{} judges
everything and errs by rejecting. Additionally, \sympyc{} accepts
off-by-one adversarial answers as a deterministic step function of
magnitude---0\% below $10^4$, 100\% at or above---because its relative
tolerance is scale-invariant. This is maximally systematic in the sense of
\citet{egashira2026delay} and invisible in any aggregate rate.

\paragraph{Contributions.}
(i) A certified metamorphic protocol for verifier auditing in which ground
truth is constructed rather than adjudicated; (ii) a per-category
decomposition across four implementations at 307{,}420 verdicts with
Wilson intervals on every cell; (iii) a contract matrix separating
implementation defects from specification ambiguity, and a coverage
measure separating rejection from execution failure; (iv) identification
of a scale-dependent false-positive mechanism with an exact threshold; and
(v) released transform suite, contract matrix, and per-sample verdict
records.

\section{Method}
\label{sec:method}

\subsection{Certified equivalence}

Let $g$ be a gold answer and $T$ a transformation. If $T$ is
semantics-preserving by construction, then $T(g)$ is correct, and a
verifier $V$ returning $V(g, T(g)) = \textsc{False}$ commits a
\textbf{certified false negative}. Dually, for a meaning-changing
transformation $T'$, acceptance is a \textbf{certified false positive}.

This design has three consequences: ground truth is free; the sample size
is combinatorial in golds $\times$ transforms $\times$ verifiers; and the
entire procedure is CPU-only.

\subsection{Transform classes and the contract matrix}

We use 43 transformations across 14 strata, partitioned into three
classes. \textsc{certified-equiv} transformations are mathematically
identical and within the verifier's declared contract; rejection is scored
as a false negative. \textsc{contract-dep} transformations depend on the
declared contract (boxing, units, text wrappers) and are reported as
specification ambiguity, never as defects. \textsc{adversarial}
transformations change meaning; acceptance is scored as a false positive.

Two design decisions matter. First, transformations producing empty output
are discarded automatically. Second, a verifier is scored only on strata
it claims to handle: a normalizer that never advertised answer extraction
is not ``wrong'' to reject a boxed answer. Out-of-contract results appear
in Table~\ref{tab:oocontract}. The contract assignments are a judgement
call and are published in the released code so that they can be contested.

\subsection{Coverage: rejection versus execution failure}
\label{sec:coverage}

A verifier can fail in two distinct ways: it can return
\textsc{False} on a correct answer, or it can fail to return a verdict at
all (parse exception, timeout, crash). Conflating these is a reporting
error with real consequences---a verifier that crashes on every input in a
stratum would otherwise be credited with a false-positive rate of
$0.0\%$, which reads as ``correctly rejected''. The same distinction
between judgement \emph{accuracy} and judgement \emph{availability} has
been argued for in security-scanner evaluation \citep{lan2026coverage}.

We therefore define the \emph{evaluated} denominator
\[
n_{\text{eval}} = n_{\textsc{True}} + n_{\textsc{False}},
\qquad
\text{coverage} = n_{\text{eval}} / n ,
\]
and report both the all-inputs rate (denominator $n$) and the evaluated
rate (denominator $n_{\text{eval}}$). Cells with zero coverage report
\textsc{n/a}, never $0$. Overall in-contract coverage is 96.5\%; the
deficit is concentrated in a single implementation
(\S\ref{sec:results}).

\subsection{Verifiers under test}

We audit four implementations: \mv{} \citep{mathverify2025} with LaTeX
extraction (library default); \mv{} restricted to plain-expression
extraction; the DeepSeek-Math lineage string normalizer
\citep{shao2024deepseekmath}; and a reference three-level cascade (exact
string, then numeric with relative tolerance $10^{-4}$, then symbolic via
SymPy). The ANTLR4 runtime is pinned to 4.13.2; behaviour differs across
runtimes.

\subsection{Data}

We use 4{,}990 unique gold answers drawn from GSM8K
\citep{cobbe2021gsm8k}, MATH across all seven subjects
\citep{hendrycks2021math}, Big-Math \citep{albalak2025bigmath}, plus
2{,}000 synthetic answers in standard MATH answer forms. Synthesis was
necessary because set and interval notation appears in only 0.1\% of
corpus answers, leaving those strata underpowered. Synthetic golds are
tagged and reported separately; \S\ref{sec:robust} confirms all findings
hold on corpus-derived answers alone.

\subsection{Execution}

The audit runs as an Azure ML parallel pipeline on five CPU nodes. Each
verification executes in an isolated subprocess with a five-second
per-item budget matching the documented symbolic-comparison timeout;
crashes and timeouts are recorded as distinct verdict classes.

\section{Results}
\label{sec:results}

%
%
%

\begin{table}[htbp]
\centering
\small
\setlength{\tabcolsep}{4pt}
\begin{tabular}{lrrrr}
\toprule
Verifier & $n$ & Self-val. & Cov. & Judged \\
\midrule
\strips{}  & 28{,}570 & \textbf{95.2\%} & 100.0\% & 95.2\% \\
\sympyc{}  & 31{,}266 & 87.3\%          & 87.3\%  & \textbf{100.0\%} \\
\mvl{}     & 31{,}266 & 79.9\%          & 100.0\% & 79.9\% \\
\mve{}     & 22{,}953 & \textbf{53.8\%} & 100.0\% & 53.8\% \\
\bottomrule
\end{tabular}
\caption{Self-validation: acceptance of certified-equivalent variants,
in-contract strata only. \emph{Self-val.}\ is over all inputs;
\emph{Cov.}\ is the fraction on which the verifier returned a verdict at
all; \emph{Judged} restricts to those. The 41.3-point spread on identical
inputs shows the figure is implementation-specific. \sympyc{} is correct
on every input it judges: its residual is entirely execution failure.}
\label{tab:selfval}
\end{table}

\begin{table}[htbp]
\centering
\footnotesize
\setlength{\tabcolsep}{3pt}
\begin{tabular}{llrr}
\toprule
Verifier & Stratum & Fail & Share \\
\midrule
\mve{}    & S6 whitespace & 7{,}852 & \textbf{74.1\%} \\
\mve{}    & S5 math-mode  & 2{,}417 & 22.8\% \\
\mvl{}    & S6 whitespace & 5{,}846 & \textbf{93.0\%} \\
\mvl{}    & S10 sets      &     238 & 3.8\% \\
\strips{} & S6 whitespace &     694 & 50.3\% \\
\strips{} & S2 frac/dec   &     438 & 31.7\% \\
\sympyc{} & S14 unreduced & 2{,}062 & \textbf{52.0\%} \\
\sympyc{} & S6 whitespace &     694 & 17.5\% \\
\bottomrule
\end{tabular}
\caption{Error mass share: fraction of each verifier's in-contract
failures attributable to each stratum, top two strata per verifier.
Whitespace and punctuation dominate both \mv{} configurations. Every
failure in the \sympyc{} rows is a parse exception rather than a rejected
answer; the full breakdown is in the released
\texttt{T2\_error\_mass\_share.csv}.}
\label{tab:errormass}
\end{table}

\begin{table}[htbp]
\centering
\footnotesize
\setlength{\tabcolsep}{3pt}
\begin{tabular}{llrrl}
\toprule
Stratum & Verif. & $n$ & FN & 95\% CI \\
\midrule
S6 white   & \mve{}    & 15{,}704 & 50.0\% & [49.2, 50.8] \\
S5 math    & \mve{}    &  4{,}951 & 48.8\% & [47.4, 50.2] \\
S10 sets   & \mvl{}    &      634 & 37.5\% & [33.9, 41.4] \\
S6 white   & \mvl{}    & 15{,}704 & 37.2\% & [36.5, 38.0] \\
S7 sqrt    & \strips{} &      404 & 35.1\% & [30.7, 39.9] \\
S2 frac    & \strips{} &  1{,}725 & 25.4\% & [23.4, 27.5] \\
S2 frac    & \mve{}    &  1{,}725 & 19.1\% & [17.3, 21.0] \\
S4 delim   & \mvl{}    &  1{,}067 & 18.7\% & [16.5, 21.2] \\
S6 white   & \strips{} & 15{,}704 &  4.4\% & [4.1, 4.8] \\
S5 math    & \strips{} &  4{,}951 &  2.1\% & [1.8, 2.6] \\
S7 sqrt    & \mvl{}    &      404 &  0.2\% & [0.0, 1.4] \\
S14 unred  & \mvl{}    &  2{,}062 &  0.0\% & [0.0, 0.2] \\
S1 frac d. & \mvl{}    &  4{,}146 &  0.0\% & [0.0, 0.1] \\
S9 group   & all       &      573 &  0.0\% & [0.0, 0.7] \\
\midrule
\multicolumn{5}{l}{\emph{zero coverage (no verdict returned)}} \\
S10 sets   & \sympyc{} &      634 & n/a & n/a \\
S14 unred  & \sympyc{} &  2{,}062 & n/a & n/a \\
\bottomrule
\end{tabular}
\caption{Certified false-negative rates by stratum and verifier,
in-contract only, with Wilson score intervals. Rates are computed over
verdicts actually returned; cells where the verifier never returned a
verdict report \textsc{n/a} rather than $0$. Strata with zero failures are
retained to show which cases are handled correctly.}
\label{tab:fnrates}
\end{table}

\begin{table}[htbp]
\centering
\small
\setlength{\tabcolsep}{4pt}
\begin{tabular}{lrrrrr}
\toprule
Mag. & $n$ & V1 & V2 & V3 & V4 \\
\midrule
$<10$     &  20 & 0.0 & 0.0 & 0.0 & 0.0 \\
$10^{1}$  & 134 & 0.0 & 0.0 & 0.0 & 0.0 \\
$10^{2}$  & 560 & 0.0 & 0.0 & 0.0 & 0.0 \\
$10^{3}$  & 367 & 0.0 & 0.0 & 0.0 & 0.0 \\
$10^{4}$  & 131 & 0.0 & 0.0 & 0.0 & \textbf{100.0} \\
$>10^{5}$ &  75 & 0.0 & 0.0 & 0.0 & \textbf{100.0} \\
\bottomrule
\end{tabular}
\caption{Off-by-one acceptance rate (\%) by gold answer magnitude, over
verdicts actually returned.
V1 $=$ \mve{}, V2 $=$ \mvl{}, V3 $=$ \strips{}, V4 $=$ \sympyc{}.
V4 steps deterministically at $10^{4}$: a relative tolerance of $10^{-4}$
is scale-invariant, so $10{,}001$ differs from $10{,}000$ by $0.01\%$ and
is accepted. Aggregate rate 16.0\% (95\% CI $[14.1, 18.1]$,
$n = 1{,}287$).}
\label{tab:offbyone}
\end{table}

\subsection{Self-validation varies by 41 points}

\begin{figure}[htbp]
\centering
\includegraphics[width=\columnwidth]{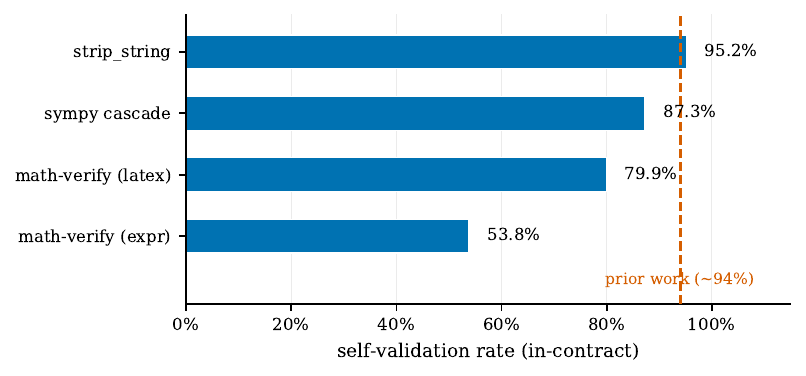}
\caption{Self-validation rate by implementation, in-contract strata. The
dashed line marks the $\sim$94\% figure reported in prior work
\citep{hfguidebook}.}
\label{fig:selfval}
\end{figure}

Table~\ref{tab:selfval} reports self-validation: the fraction of
certified-equivalent variants a verifier accepts. The range is 53.8\% to
95.2\% on identical inputs. The published $\sim$94\% figure
\citep{hfguidebook} falls near the top of this range rather than at its
centre, indicating that it characterises one particular harness on one
particular input distribution rather than the task.
Figure~\ref{fig:selfval} plots the spread.

The coverage column separates two failure modes that the aggregate hides.
\sympyc{} returns a verdict on 87.3\% of inputs and is correct on
\emph{100\%} of those it judges: its entire residual is parse failure, not
equivalence error. The two \mv{} configurations return verdicts on 100\%
of inputs and err by rejecting. These are different defects requiring
different fixes, and an aggregate rate conflates them.

\subsection{Whitespace dominates the error budget}

\begin{figure}[htbp]
\centering
\includegraphics[width=\columnwidth]{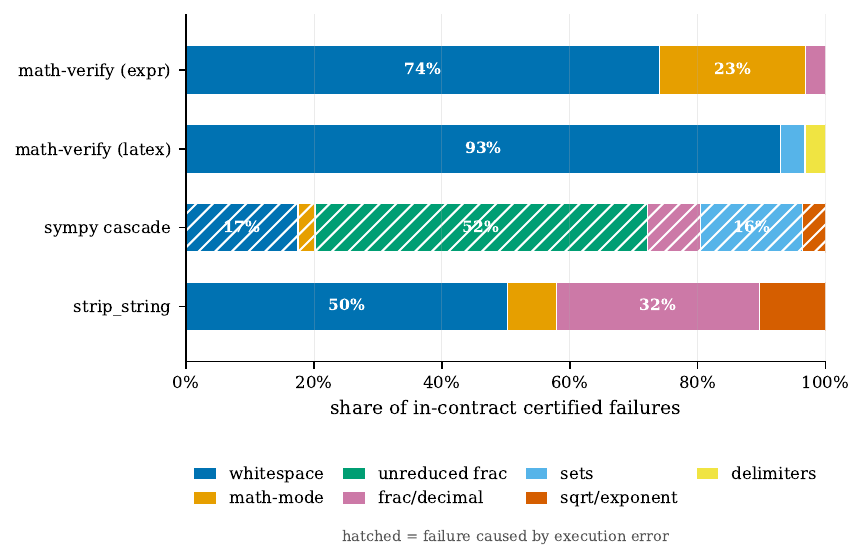}
\caption{Error budget: share of in-contract certified failures by stratum.
Hatched segments indicate failures caused by execution error rather than
rejection.}
\label{fig:budget}
\end{figure}

Table~\ref{tab:errormass} and Figure~\ref{fig:budget} decompose
in-contract failures by stratum. For \mvl{}, whitespace and punctuation
account for 93.0\% of all failures; for \mve{}, 74.1\%. The remaining mass
is spread thinly across sets, delimiters, and fraction handling.

Table~\ref{tab:fnrates} gives per-stratum rates with Wilson intervals. The
whitespace stratum is measured at $n = 15{,}704$ with intervals of
$\pm 0.8$ points, so the effect is not a small-sample artefact.
Table~\ref{tab:examples} shows representative pairs: an expression is
accepted, and the same expression followed by a period is rejected.

\subsection{A scale-dependent false-positive mechanism}

\begin{figure}[htbp]
\centering
\includegraphics[width=\columnwidth]{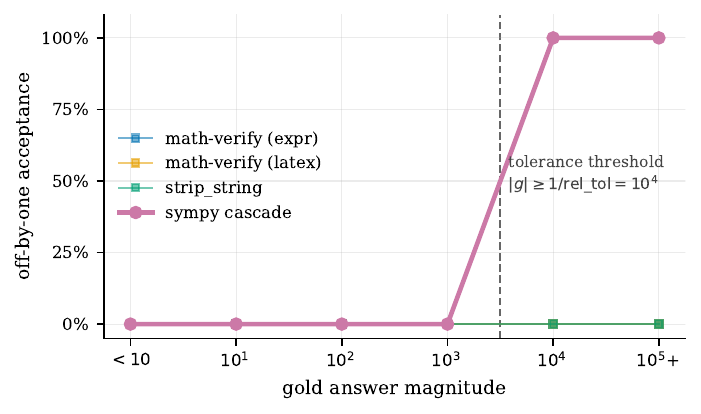}
\caption{Off-by-one acceptance by gold answer magnitude. The reference
cascade steps deterministically at $10^{4}$; the other three verifiers
reject all off-by-one variants at every magnitude.}
\label{fig:offbyone}
\end{figure}

Table~\ref{tab:offbyone} and Figure~\ref{fig:offbyone} report acceptance
of off-by-one adversarial answers by gold magnitude. Three verifiers
reject all of them at every magnitude. \sympyc{} rejects all of them below
$10^4$ and accepts \emph{all} of them at or above $10^4$. The mechanism is
that a relative tolerance of $10^{-4}$ is scale-invariant: for a gold
answer of $10{,}000$, the value $10{,}001$ differs by $0.01\%$ and is
accepted. The aggregate off-by-one false-positive rate for this verifier
is 16.0\% (95\% CI $[14.1, 18.1]$, $n = 1{,}287$), but the aggregate
conceals a threshold.

\subsection{Reported accuracy is not comparable across papers}

Table~\ref{tab:contract} shows acceptance on contract-dependent input
classes. Acceptance on boxed answers ranges from 0.0\% to 75.1\%; on
scientific notation, from 0.0\% to 100.0\%. These are not defects---each
implementation behaves as designed---but the differences are undocumented.

Table~\ref{tab:disagree} quantifies the consequence. Two configurations of
the same library disagree on 49.9\% of certified-equivalent pairs; panel
(b) of the same table gives the pair counts supporting each cell.

\subsection{Robustness: corpus-only}
\label{sec:robust}

Excluding all synthetic golds, self-validation is 57.5\% (\mve{}), 84.1\%
(\mvl{}), 87.8\% (\strips{}), and 88.3\% (\sympyc{}). Ordering and
magnitude are preserved; synthesis is not driving the results.

\section{Discussion}

\paragraph{The residual is string hygiene, not grammar.}
Prior attribution---``LaTeX parsing failures''---implies difficult grammar
work. The decomposition shows otherwise: 93.0\% of one verifier's
in-contract failures are whitespace and punctuation. This is more
actionable and more surprising than the aggregate suggests.

\paragraph{Verifier configuration must be reported.}
Acceptance on identical inputs ranges 0\%--100\% by configuration, and two
configurations of one library disagree on 49.9\% of cases. A reported
benchmark accuracy is therefore a function of the verifier configuration,
which is almost never stated. We recommend reporting verifier identity,
version, extraction configuration, and runtime.

\paragraph{Coverage is a first-class metric.}
\S\ref{sec:coverage} is not merely a reporting nicety. Two verifiers with
similar aggregate residuals---\sympyc{} at 12.7\% and \strips{} at
4.8\%---fail for opposite reasons, and only the coverage split reveals it.
We recommend that verifier evaluations report coverage alongside accuracy.

\paragraph{Scale-dependent false positives.}
Table~\ref{tab:offbyone} shows a threshold rather than a rate.
\citet{egashira2026delay} distinguish systematic from random verifier
error and show that systematic false positives can produce plateaus or
collapse, with outcomes determined by the error pattern rather than the
overall rate. A magnitude threshold is maximally systematic and invisible
in any aggregate figure.

\paragraph{Relation to reported false-negative rates.}
Reported false-negative rates near 38\% \citep{cai2025noisy} are measured
on real model outputs, which include mathematical non-equivalences such as
unreduced fractions. Our measurement isolates the formatting and
equivalence component under certified transformations and is therefore a
lower bound on that quantity. The two are complementary.

\section{Limitations}

\textbf{Verifier-level, not model-level.} We do not estimate the effect on
any model's benchmark score; real outputs do not produce transform
variants at equal rates. \textbf{No training runs.} We do not measure
downstream RLVR effect; \citet{zhang2026leaky} examine hardened versus
leaky rewards in code RLVR and find bounded effects. \textbf{Open
implementations only.} No claims about closed frontier systems.
\textbf{Synthetic component.} 2{,}000 of 4{,}990 golds are synthetic;
\S\ref{sec:robust} reports corpus-only results separately.
\textbf{Contract assignment is a judgement.} We publish the matrix so it
can be contested.

\section{Related Work}

\paragraph{Verifier noise in RLVR.}
\citet{cai2025noisy} model the verifier as a stochastic reward channel
with asymmetric noise rates and derive corrections requiring the
false-negative rate as input. Their motivating example, a checker marking
$\tfrac{12}{36}$ wrong against a canonical $\tfrac{1}{3}$, is one of our
strata. Their method assumes the rates; we measure and decompose them.
\citet{egashira2026delay} distinguish systematic from random verifier
error and show outcomes are determined by the error pattern; our
per-stratum decomposition characterises that pattern, and the magnitude
threshold in \S\ref{sec:results} is a maximally systematic instance.
\citet{huang2025accuracy} compare rule-based and model-based verifiers and
find opposing failure directions; we extend the false-negative side with a
category-level decomposition and add certified adversarial probes.

\paragraph{Verifier and benchmark auditing.}
\citet{ammanamanchi2026faults} audit five Lean theorem-proving benchmarks
with corpus-scale static checkers, surfacing 4{,}833 findings and
proposing a fault taxonomy with released tooling. We adopt the same
posture for natural-language answer verification. In code RLVR,
\citet{zhang2026leaky} run a preregistered causal contrast between leaky
and hardened reward suites, and \citet{rajan2026hackability} measure that
28.5\% of SWE-bench Verified tasks accept Docker-verified incorrect
patches. Outside RLVR, \citet{lan2026coverage} make a structurally similar
argument for security scanners: conventional metrics characterise only
cases where a tool yields a usable judgement, so coverage must be reported
separately from accuracy.

\paragraph{Metamorphic testing.}
Semantics-preserving transformation as a robustness probe has an
established lineage \citep{asgari2025mtsurvey,hyun2024metal,zhou2026lgmt}.
All of this work perturbs the \emph{input} to test the \emph{model}. We
invert the target: the transformation is applied to the answer string, and
the system under test is the grader.

\paragraph{Evaluation brittleness.}
\citet{su2025delimiter} show that changing the single character separating
in-context examples moves MMLU accuracy by up to $\pm 23\%$.
\citet{hua2025flaw} argue much reported prompt sensitivity is an artefact
of heuristic scoring. Our results are the verifier-side counterpart to
both. \citet{miller2024errorbars} argues for statistical rigour in
evaluation reporting; we attach Wilson intervals to every cell.

\section{Conclusion}

Verifier reliability is a measurable property that varies by 41 points
across widely used implementations, is dominated by whitespace handling
rather than mathematical parsing, and includes at least one deterministic
false-positive mechanism triggered by answer magnitude. Separating
rejection from execution failure shows that implementations with similar
aggregate residuals fail for opposite reasons. The transform suite,
contract matrix, and per-sample verdict records are released with this
paper.

\bibliography{references}

\appendix

\section{Supplementary Tables}
\label{sec:apptables}

Tables~\ref{tab:disagree}--\ref{tab:examples} give the cross-verifier
disagreement matrix with its pair support, acceptance on
contract-dependent input classes, out-of-contract behaviour, and
representative certified-equivalent pairs that production verifiers
reject. Throughout, V1 $=$ \mve{}, V2 $=$ \mvl{}, V3 $=$ \strips{}, and
V4 $=$ \sympyc{}.

%
%

\begin{table}[htbp]
\centering
\small
\setlength{\tabcolsep}{4pt}
\begin{tabular}{lrrrr}
\toprule
\multicolumn{5}{l}{\emph{(a) Disagreement rate (\%)}} \\
\midrule
   & V1   & V2   & V3   & V4   \\
V1 & ---  & \textbf{49.9} & 48.0 & 53.4 \\
V2 & 49.9 & ---  & 31.4 & 22.0 \\
V3 & 48.0 & 31.4 & ---  & 0.4  \\
V4 & 53.4 & 22.0 & 0.4  & ---  \\
\midrule
\multicolumn{5}{l}{\emph{(b) Pairs judged by both}} \\
\midrule
   & V1     & V2     & V3     & V4     \\
V1 & 31{,}266 & 31{,}266 & 31{,}266 & 27{,}299 \\
V2 & 31{,}266 & 31{,}266 & 31{,}266 & 27{,}299 \\
V3 & 31{,}266 & 31{,}266 & 31{,}266 & 27{,}299 \\
V4 & 27{,}299 & 27{,}299 & 27{,}299 & 27{,}299 \\
\bottomrule
\end{tabular}
\caption{Pairwise disagreement on certified-equivalent pairs, computed
only over pairs where both verifiers returned a verdict.
V1 $=$ \mve{}, V2 $=$ \mvl{}, V3 $=$ \strips{}, V4 $=$ \sympyc{}.
Two configurations of the same library (V1, V2) disagree on 49.9\% of
cases. Support is lower for pairs involving V4 because it does not return
a verdict on every input.}
\label{tab:disagree}
\end{table}

\begin{table}[htbp]
\centering
\small
\setlength{\tabcolsep}{4pt}
\begin{tabular}{lrrrr}
\toprule
Input class & $n$ & V1 & V2 & V3 \\
\midrule
boxed       & 19{,}952 &  8.0 & \textbf{75.1}  &   0.0 \\
scientific  &      326 &  0.0 & \textbf{100.0} &   0.0 \\
with unit   &  2{,}782 & 50.0 & \textbf{100.0} & 100.0 \\
text prefix &  9{,}980 & 28.2 & 39.6           &  48.9 \\
percent     &       20 &  0.0 & \textbf{100.0} &   0.0 \\
\bottomrule
\end{tabular}
\caption{Acceptance rate (\%) on contract-dependent input classes, over
verdicts actually returned.
V1 $=$ \mve{}, V2 $=$ \mvl{}, V3 $=$ \strips{}.
Classes are a boxed answer, scientific notation, a trailing unit, the
prefix ``The answer is'', and a percent sign. Acceptance spans 0--100\%
on identical inputs: these are undocumented contract differences, not
defects. \sympyc{} is omitted because it returns no verdict on boxed or
scientific inputs.}
\label{tab:contract}
\end{table}

\begin{table}[htbp]
\centering
\small
\setlength{\tabcolsep}{4pt}
\begin{tabular}{llrr}
\toprule
Stratum & Verifier & $n$ & Rejection \\
\midrule
S10 sets   & \strips{} &      634 & 100.0\% \\
S14 unred  & \mve{}    &  2{,}062 & 100.0\% \\
S14 unred  & \strips{} &  2{,}062 & 100.0\% \\
S10 sets   & \mve{}    &      634 &  97.5\% \\
S1 frac d. & \mve{}    &  4{,}146 &  93.8\% \\
S4 delim   & \mve{}    &  1{,}067 &  87.1\% \\
S7 sqrt    & \mve{}    &      404 &  31.4\% \\
\bottomrule
\end{tabular}
\caption{Out-of-contract behaviour: strata a verifier does not claim to
handle. Reported for completeness; not counted as defects.}
\label{tab:oocontract}
\end{table}

\begin{table}[htbp]
\centering
\footnotesize
\setlength{\tabcolsep}{4pt}
\begin{tabular}{lll}
\toprule
Gold & Variant & Rejected by \\
\midrule
\verb|\frac{1}{2}|  & \verb|\frac{1}{2}.|    & V2, V1 \\
\verb|\frac{1}{2}|  & \verb|\frac{1}{2}\n|   & V2, V1 \\
\verb|\frac{1}{2}|  & \verb|\dfrac{1}{2}|    & V1 \\
\verb|(a+5)(b+2)|   & \verb|\left(a+5\right)| & V1 \\
\verb|3\pm\sqrt{2}| & \verb|3\,\pm\,\sqrt{2}| & V3 \\
\bottomrule
\end{tabular}
\caption{Certified-equivalent pairs rejected by production verifiers.
V1 $=$ \mve{}, V2 $=$ \mvl{}, V3 $=$ \strips{}. A trailing period changes
the verdict. The fourth variant is truncated for width; the full form is
\texttt{$\backslash$left(a+5$\backslash$right)$\backslash$left(b+2$\backslash$right)}.}
\label{tab:examples}
\end{table}

\end{document}